# A Safety-Gated Multimodal AI Backend for Mental-Health Support: Hierarchical State Representation, Conservative Risk Fusion, and Controlled Generation in Anian

Lei Wang[1]; Xiao Wang[1]; Lei Li[2]

1 RYTECH, Wuhan, 430074, China; 2 City University of Hong Kong, Hong Kong SAR, China

## Abstract

Safety-critical mental-health support systems require architectures that can identify when ordinary supportive conversation is appropriate and when free-form generation should be blocked. This paper presents Anian, a safety-gated multimodal AI backend designed for perinatal mental-health support and mindfulness-intervention routing. Anian is not intended to diagnose depression, anxiety, bipolar disorder, postpartum psychosis, suicide risk, or any other psychiatric condition, and it is not a substitute for clinical care or crisis intervention. The technical contribution is a modular pipeline that places generative AI downstream of structured state representation, conservative risk fusion, and response gating. The system maps user text or voice-derived ASR transcripts into four linked layers: L1 emotion states, L2 psychosocial constructs, L3 safety risk, and L4 intervention routes. Local text and rule-based safety evidence are fused with external voice-derived safety evidence by a highest-risk-priority rule, S_fusion = max(S_local, S_external). If the fused safety level reaches moderate or high risk, ordinary AI-generated responses and ordinary text-to-speech delivery are blocked and replaced by fixed safety-oriented content and human-support prompts. An internal prototype technical evaluation was conducted using approximately 858,295 normalized records from public emotion, dialogue, mental-health-related, and Chinese dialogue corpora under a weak-label and rule-derived evaluation framework. The L1 emotion model achieved a micro-F1 of 0.9604, the L2 psychosocial-construct model achieved a micro-F1 of 0.9144, and the L4 routing model achieved a micro-F1 of 0.9742. In a controlled safety stress test of 233 samples, the L3 rule engine achieved high-risk recall of 1.0000 within the predefined test scenarios. These results demonstrate internal feasibility of the current label framework and gating logic, but they do not establish clinical validity, diagnostic accuracy, real-world safety, or clinical effectiveness. The paper reports the architecture, ontology, safety-fusion mechanism, prototype evaluation, error-analysis plan, and validation roadmap for future expert-reviewed and real-world evaluation.

## 1 Introduction

Generative AI systems are increasingly capable of producing fluent, empathic, and context-sensitive responses. In mental-health support contexts, however, fluency is not sufficient. A system may be conversationally persuasive while still failing to detect self-harm expressions, infant- or others-harm concerns, acute loss of control, or ambiguous safety signals. For safety-sensitive support tools, the central technical question is therefore not only how to generate a helpful reply, but also how to determine when ordinary generation should not occur.

Perinatal mental health provides a demanding application context for this problem. Clinical guidance emphasizes that perinatal care should consider depression, anxiety and related disorders, bipolar disorder, postpartum psychosis, suicidality, and broader safety concerns, and that screening must connect to assessment, treatment, follow-up, and safety response pathways [1-4]. Standard instruments such as the Edinburgh Postnatal Depression Scale, Patient Health Questionnaire-9, and Generalized Anxiety Disorder-7 scale are useful for population screening and initial symptom assessment [5-7], but everyday user language often expresses dynamic mixtures of worry, sleep disruption, self-blame, shame, relationship stress, social isolation, caregiving pressure, and help-seeking intent rather than a single diagnostic category [8,9].

Digital interventions and mobile mental-health applications can expand access to low-intensity support. Recent systematic reviews and meta-analyses suggest that digital and app-based interventions may reduce symptoms of depression, anxiety, or stress in selected populations, while also noting heterogeneity in design, adherence, follow-up duration, personalization, and workflow integration [10-14]. Large language models and conversational agents may further support psychoeducation, emotional support, screening assistance, and digital intervention delivery, but reviews consistently identify unresolved issues in reliability, evaluation standards, expert annotation, interpretability, privacy, and ethical governance [15-18].

These concerns motivate a safety-gated architecture rather than an LLM-first architecture. In Anian, user input is not passed directly to an open-ended generator as the primary decision-maker. Instead, the system first constructs a structured state representation, performs conservative safety assessment, selects an intervention route, and only then determines whether ordinary generative output is allowed. This reframes personalization as an action-selection and safety-orchestration problem rather than only a language-generation problem.

This paper makes four technical contributions. First, it defines a hierarchical L1-L4 state representation that separates affective expression, intervention-relevant psychosocial constructs, safety risk, and support routing. Second, it implements a hybrid model-rule safety layer with explicit handling of direct safety triggers and common false-positive contexts. Third, it introduces a conservative multimodal safety-fusion and generation-gating mechanism in which the highest-risk signal across modalities controls downstream generation. Fourth, it reports an internal prototype technical evaluation and a staged validation roadmap for human gold-standard annotation, safety review, ASR robustness testing, ablation analysis, and real-world workflow evaluation.

## 2 Related work

### 2.1 AI and NLP for mental-health support

AI has been used in mental-health contexts for text classification, risk prediction, social-media signal detection, conversational support, and digital intervention assistance [15-18,49-52]. In perinatal mental health, recent work suggests potential use cases for early identification, care-cascade optimization, and risk stratification, but the field remains limited by under-representative data, insufficient external validation, heterogeneous outcomes, uncertain clinical implementability, and potential bias [27-30].

From a computational perspective, mental-health support differs from ordinary dialogue modeling because a small number of safety-critical cases can dominate the harm profile. Standard aggregate metrics such as accuracy or micro-F1 may overstate usefulness when high-risk classes are rare. A deployable system therefore requires dedicated safety endpoints, including high-risk recall, false-negative rate, false-positive burden, escalation appropriateness, and generation leakage after a safety block.

### 2.2 Generative AI safety and response control

LLMs can produce emotionally responsive and coherent language, but they can also provide inappropriate reassurance, hallucinate, anthropomorphize support, or fail to follow escalation boundaries in safety-critical situations [15-20]. WHO guidance on AI for health and large multimodal models emphasizes safety, transparency, human oversight, accountability, and governance rather than capability alone [19,20].

Anian addresses this problem by treating the generative model as a constrained expression component. The system does not ask the generator to be the primary safety adjudicator. Instead, generation is made conditional on an upstream risk-fusion decision. When fused risk is moderate or high, ordinary generation is unavailable by design.

### 2.3 Adaptive intervention routing and mindfulness content

Digital therapeutics, mindfulness applications, and low-intensity psychological support tools often organize content as predefined modules. Meta-analytic evidence suggests potential short-term effects of mental-health smartphone applications and mindfulness applications, but individual responses vary, engagement is difficult to sustain, and long-term outcomes remain uncertain [12-14].

The design insight behind Anian is that intervention routing should be state-dependent and safety-aware. Worry may map to breathing or grounding; sleep distress may map to a brief sleep-oriented practice; self-blame and shame may map to self-compassion; help-seeking and isolation may map to support prompts; and moderate or high safety risk should override ordinary content routing altogether. This makes personalization a route-selection problem with explicit safety constraints.

# 3 System design and methods

## 3.1 Study design, intended use, and non-use

This is a system-development and internal prototype technical-evaluation study. Anian is intended as a backend component for perinatal psychological support, mindfulness-course routing, and safety-aware response control. It is not intended to diagnose depression, anxiety, bipolar disorder, postpartum psychosis, suicide risk, or other psychiatric disorders. It is not a medical device claim, clinical screening conclusion, treatment recommendation, or crisis-intervention service in the current evidence stage.

The present evaluation reports architecture, label ontology, model-rule integration, speech-input handling, conservative safety fusion, generation gating, prototype-level performance, error-analysis design, and a staged validation roadmap. Because no prospective real-user clinical evaluation has been completed, no clinical effectiveness outcome is reported.

## 3.2 Overall architecture

Anian consists of an input layer, speech-analysis interface, local state-prediction layer, safety-fusion layer, intervention-routing layer, and generation-control layer. The input layer accepts text and voice. Voice input is processed by a speech-analysis service that returns ASR transcription, tone/prosody information, an external safety signal, and optional TTS capability. The ASR transcript enters the local Anian pipeline.

The local pipeline is executed in a safety-first order: L3 safety rules, L1 emotion recognition, L2 psychosocial construct recognition, and L4 intervention routing. Local safety evidence and external voice-derived safety evidence are then fused by a conservative highest-risk-priority rule. The fused risk level determines frontend actions and whether ordinary AI-generated output is permitted.

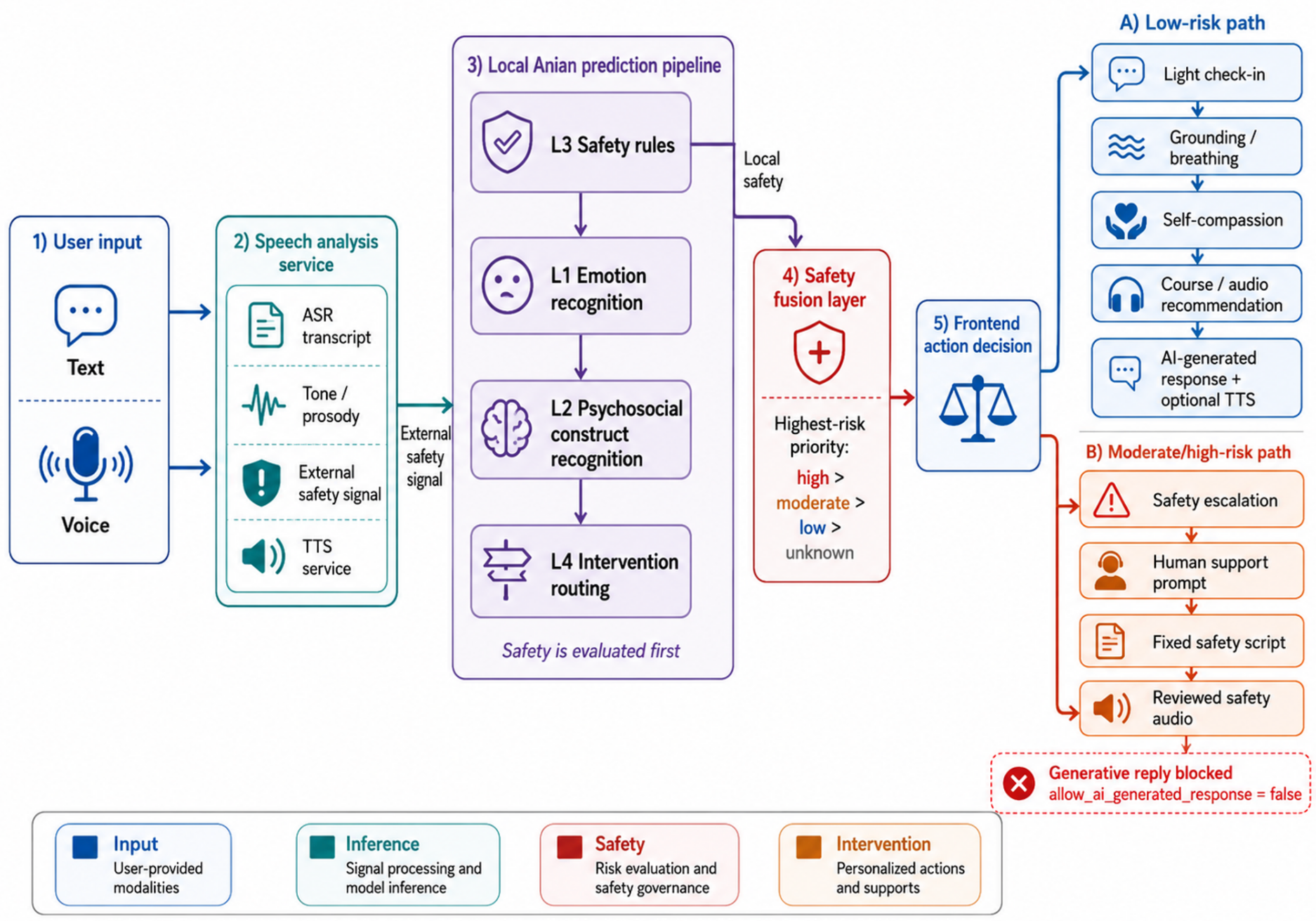

**Figure 1 |** Overall architecture of the Anian safety-gated multimodal backend. Text and voice input are transformed into local state predictions and external speech-derived safety evidence. The safety-fusion layer preserves the highest-risk signal across modalities.

Low-risk states may receive ordinary support routes, whereas moderate- or high-risk states trigger safety escalation and block ordinary AI-generated response and ordinary TTS delivery.

### 3.3 Hierarchical L1-L4 state representation

Anian decomposes user expression into four operational layers. L1 identifies surface affective states, including sadness, anxiety_fear, anger_irritability, guilt_shame, calm_relief, joy_positive, and neutral. L2 identifies intervention-relevant psychosocial constructs, including worry, sleep_distress, self_blame, relationship_stress, caregiving_stress, isolation, emotional_overload, help_seeking, settling, and acute_or_safety_distress. L3 assigns a safety level of none, low, moderate, or high. L4 maps the state representation to an actionable support route, such as light_check_in, breathing_grounding, downshift_grounding, sleep_grounding_short_practice, self_compassion, support_seeking_prompt, ask_support_or_professional_path, or safety_escalation.

This representation is designed for support routing and safety control rather than diagnosis. Labels such as sadness, self_blame, or sleep_distress denote expressed states or themes in user language. They do not imply psychiatric diagnoses.

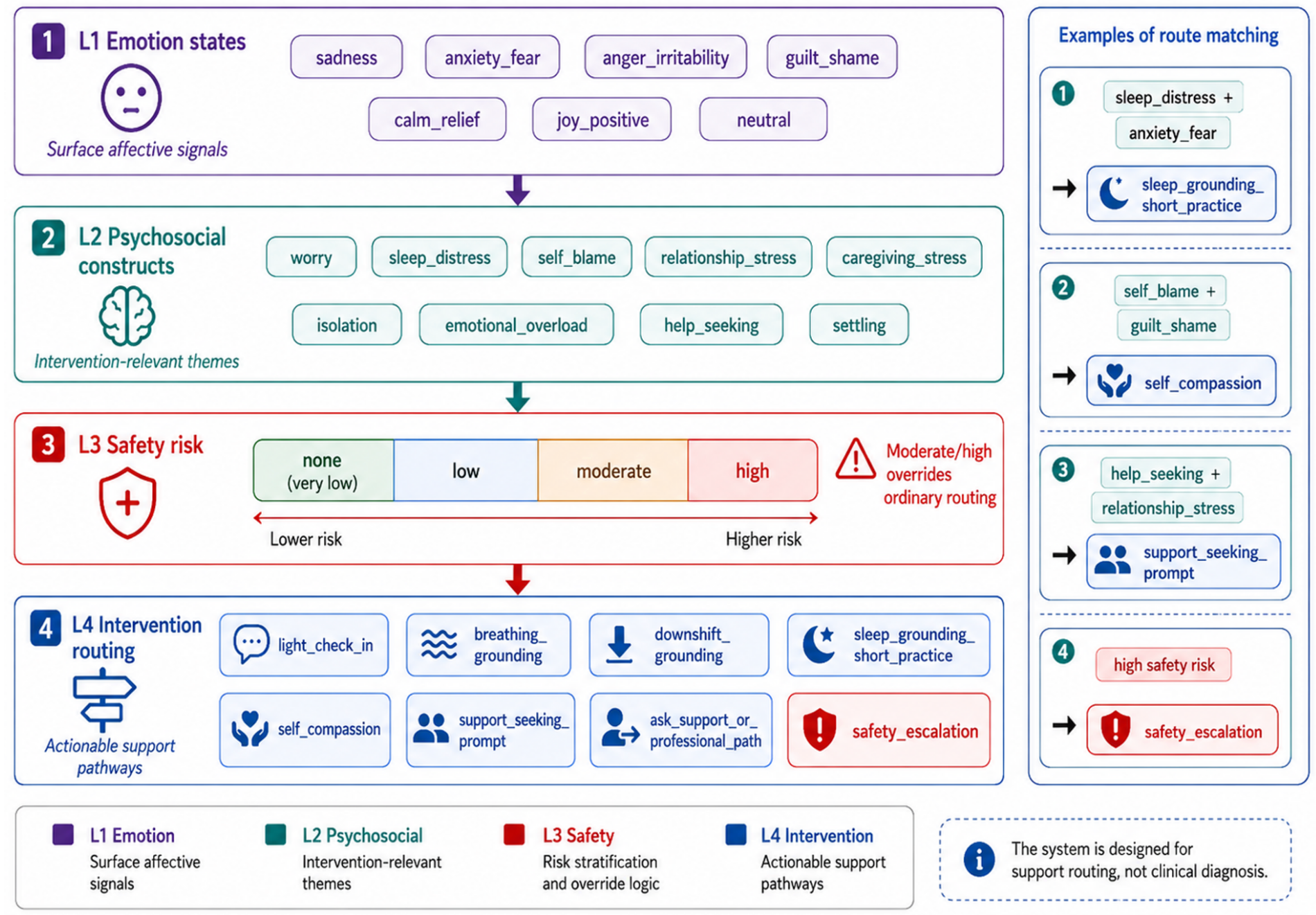

**Figure 2** | Hierarchical L1-L4 state representation and intervention routing. L1 captures affective states; L2 captures intervention-relevant psychosocial constructs; L3 stratifies safety risk; and L4 selects a support route. Moderate and high L3 states override ordinary routing.

**Table 1** | Anian L1-L4 state representation.

| Layer | Task | Representative labels | Output use | Safety implication |
|---|---|---|---|---|
| L1 | Emotion-state recognition | sadness; anxiety_fear; guilt_shame | Surface affective characterization | Does not directly determine safety level |
| L2 | Psychosocial construct recognition | worry; sleep_distress; self_blame | Content and practice matching | Signals intervention-relevant distress themes |
| L3 | Safety-risk stratification | none; low; moderate; high | Risk handling and override logic | Moderate/high risk overrides ordinary routing |
| L4 | Intervention routing | breathing_grounding; self_compassion; safety_escalation | Frontend action selection | Safety escalation replaces ordinary support |

### 3.4 Safety rules, conservative fusion, and generation gating

The L3 safety layer identifies expressions of suicidal ideation, self-harm, harm to an infant or others, and acute loss of control. It also includes false-positive suppression logic for non-risk expressions such as colloquial exaggerations, negated statements, quotations, third-party discussion, and commercial idioms. If L3 returns moderate or high risk, ordinary L4 routing is overridden and the final route is forced to safety_escalation. In later expert review, L3 operational definitions can be compared against established suicide-risk instruments such as the Columbia-Suicide Severity Rating Scale [31].

Safety fusion is defined as S_fusion = max(S_local, S_external), where S_local is generated by the local text model and L3 rules, and S_external is derived from the speech-analysis service. The risk order is high > moderate > low > none > unknown. This rule intentionally prevents a calm voice tone, low-risk prosody, or unavailable external signal from downgrading high-risk textual content.

**Algorithm 1 | Conservative safety fusion and generation gating**

Input: user text or ASR transcript x; optional external speech-safety output e; local L3 safety output l.
1. Compute local safety evidence S_local from L3 rules, text-state outputs, negation/context filters, and explicit safety triggers.
2. Compute external safety evidence S_external from speech-analysis service outputs when available; otherwise set S_external = unknown.
3. Define the risk order as high > moderate > low > none > unknown.
4. Set S_fusion = max(S_local, S_external) under the risk order.
5. If S_fusion is moderate or high, set allow_ai_generated_response = false, block ordinary TTS, force route = safety_escalation, and return fixed safety content plus human-support prompts.
6. If S_fusion is low, none, or unknown, permit ordinary state-based routing subject to route-specific safeguards and logging.
Output: fused risk level, route, generation-gate decision, and auditable safety evidence.

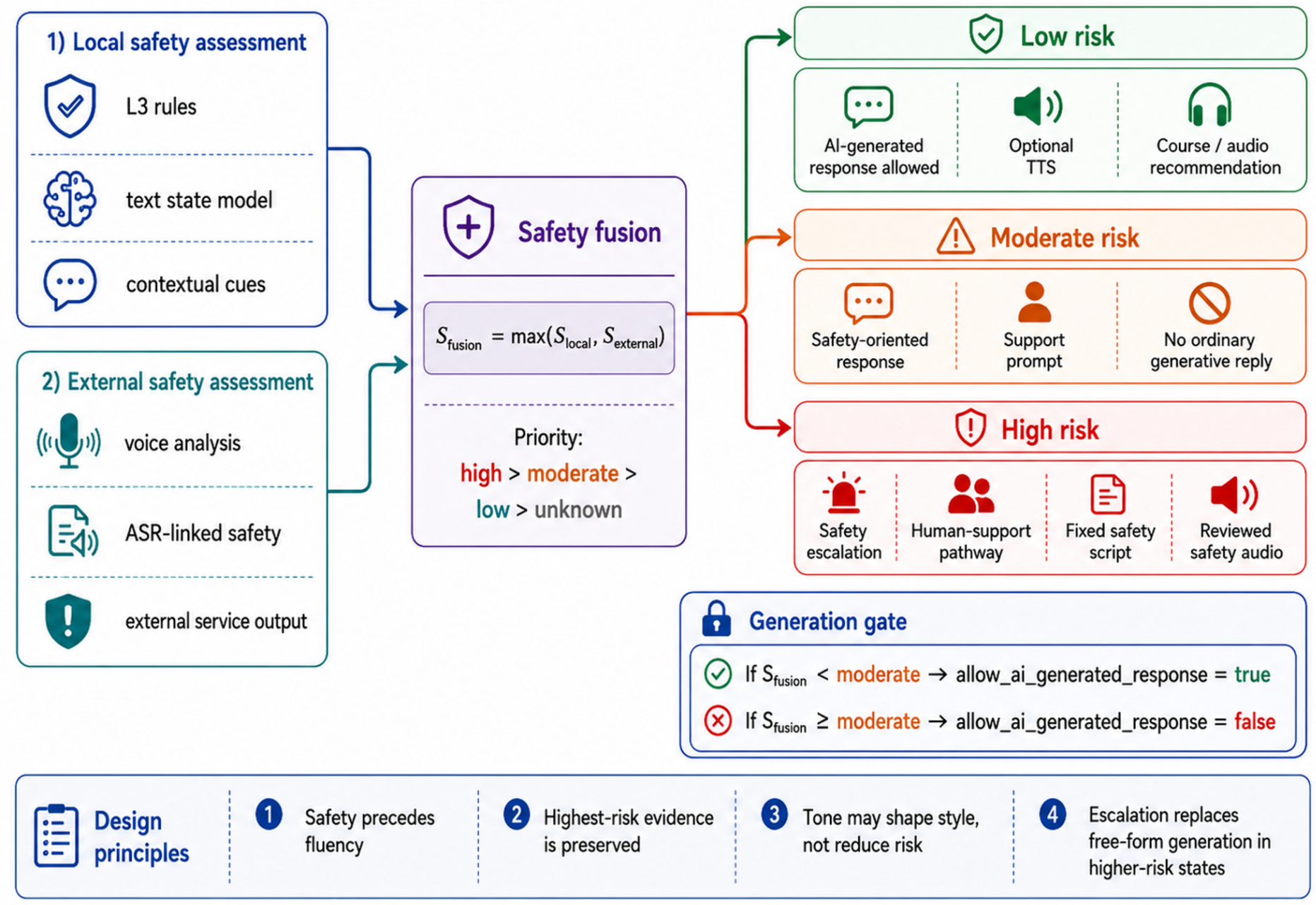

**Figure 3** | Safety fusion and generation gating. Local safety assessment integrates L3 rules, text-state model outputs, and contextual filters. External safety assessment integrates speech-analysis outputs. The generation gate permits ordinary AI-generated responses only when fused risk remains below the moderate threshold.

### 3.5 Data sources, label provenance, and evidence boundaries

The current corpus consists of approximately 858,295 normalized records from public emotion, dialogue, mental-health-related, and Chinese dialogue datasets according to the internal Anian evidence statement [32]. Sources include GoEmotions, DailyDialog, EmpatheticDialogues, MELD, EmoWOZ, CPED, M3ED, and CAMS [33-39]. These data are suitable for internal prototype experiments in emotion understanding, distress-signal detection, safety triage, and course routing. They are not perinatal clinical interview data and do not constitute a diagnostic gold standard.

Labels in the current evaluation originate from three sources: original labels in public datasets, Anian rule-derived mappings, and AI/rule-assisted automatic labels. A human-review specification and gold-set design have been prepared, but a traceable independent human gold-standard annotation loop has not yet been completed. Accordingly, the reported performance should be interpreted as weak-label prototype technical evaluation rather than clinical validation or real-world safety certification.

For reproducibility, a final public release should report source-level counts, split counts, label-source proportions, thresholds, random seeds, and preprocessing hashes from the frozen manifest. The present arXiv version reports the available aggregate evidence boundary and makes this limitation explicit rather than presenting the corpus as a clinical gold-standard dataset.

**Table 2 |** Data sources and evidence boundaries.

| Data source | Language/type | Role in Anian | Main limitation |
|---|---|---|---|
| GoEmotions | English social-media emotion | L1 emotion mapping and pretraining reference | Non-perinatal; English social-media context |
| DailyDialog / EmpatheticDialogues | English daily and empathic dialogue | Dialogue emotion and support-expression learning | Not clinical interviews |
| MELD / EmoWOZ | English multimodal or task-oriented dialogue | Dialogue-emotion representation | Performed or task-oriented context |
| CPED / M3ED | Chinese emotional dialogue | Chinese expression adaptation | Not perinatal clinical gold standard |
| CAMS | English mental-health-related posts | Safety-signal exploration | Social-media context; not perinatal cohort |
| Anian merged corpus | Multi-source normalized corpus | Prototype training and evaluation | Weak-label and rule-derived evidence |

### 3.6 Models and evaluation metrics

The L1 and L2 modules use Transformer/BERT-family Chinese text-classification models for multi-label prediction. The L4 module is a single-label route classifier. L3 is implemented primarily as a rule engine. Transformers and BERT provide the foundation for contextual text representation [40,41].

L1 and L2 multi-label tasks were evaluated using precision, recall, F1-score, micro-F1, macro-F1, weighted-F1, exact match accuracy, and hamming loss. L4 routing was evaluated using accuracy, micro-F1, macro-F1, and weighted-F1. L3 safety evaluation prioritized high-risk recall, false-negative rate, false-positive rate, safety-escalation recall, and confusion matrices. For safety tasks, recall of moderate/high-risk expressions is more important than overall accuracy because false negatives may carry greater harm than false positives in this setting.

In the planned formal validation phase, bootstrap resampling will estimate 95% confidence intervals for primary performance metrics, and paired comparisons among text-only, rule-only, text-plus-rule, voice-safety-only, and full fusion strategies may use McNemar tests for paired binary decisions [42,43]. For highly imbalanced high-risk data, PR-AUC should be reported together with recall-oriented safety metrics. Classification metrics should be interpreted cautiously in clinical contexts and not treated as proxies for safety or utility without decision-specific evaluation [48].

## 4 Prototype technical evaluation

### 4.1 Interpretation of prototype metrics

The results below evaluate whether the current system learned and executed the current weak-label framework and safety-gating logic. They do not establish clinical validity of L1-L4 constructs, diagnostic performance, real-world safety, user acceptability, or clinical effectiveness.

### 4.2 Model performance under the current weak-label framework

The current default L1 emotion model achieved a micro-F1 of 0.9604, macro-F1 of 0.6831, weighted-F1 of 0.9671, exact match accuracy of 0.9515, and hamming loss of 0.0100 on its internal test set. The L2 psychosocial-construct model achieved a micro-F1 of 0.9144, macro-F1 of 0.8055, weighted-F1 of 0.9212, exact match accuracy of 0.9104, and hamming loss of 0.0095. The L4 intervention-routing model achieved accuracy/micro-F1 of 0.9742, macro-F1 of 0.8351, and weighted-F1 of 0.9746.

These metrics indicate internal learning of the current label and routing policy, not clinical construct validity. They should therefore be used to motivate the next validation stage rather than to claim deployment readiness.

**Table 3 |** Current prototype-level model performance.

| Module | Task | Model/strategy | Primary metric | Current result | Evidence boundary |
| --- | --- | --- | --- | --- | --- |
| L1 | Emotion-state recognition | Chinese BERT multi-label model | micro-F1 | 0.9604 | Prototype-level emotion-label learning |
| L2 | Psychosocial construct recognition | Chinese BERT multi-label model | micro-F1 | 0.9144 | Weak-label construct learning |
| L3 | Safety-risk classification | Rule engine | high-risk recall | 1.0000 | Controlled stress-test only |
| L4 | Intervention routing | Chinese BERT route classifier | micro-F1 | 0.9742 | Routing-policy learning, not clinical efficacy |

### 4.3 Controlled safety stress testing

In a controlled safety stress test of 233 samples, the L3 safety rules achieved safety accuracy, high-risk recall, safety-escalation recall, and false-positive flag recall of 1.0000, with zero mismatches in the predefined scenarios. The stress test included direct self-harm or suicidal expressions, infant/others-harm expressions, acute loss-of-control expressions, and common Chinese false-positive expressions.

This result indicates coverage of the designed stress-test scenarios. It must not be interpreted as real-world zero false negatives. Real-world testing requires broader language variation, independent expert gold labels, ASR perturbation, dialect/accent testing, multi-turn accumulation, and deployment-condition evaluation.

**Table 4 |** L3 safety stress-test results.

| Metric | Result | Interpretation | Limitation |
| --- | --- | --- | --- |
| Sample size | 233 | Controlled safety stress test | Not a real-world cohort |
| Safety accuracy | 1.0000 | Expected safety label matched | Dependent on predefined scenarios |
| High-risk recall | 1.0000 | No high-risk misses in the test set | Does not imply real-world zero misses |
| Safety escalation recall | 1.0000 | Escalation triggered when expected | Requires expert-reviewed gold set |
| False-positive flag recall | 1.0000 | False-positive suppression behaved as expected | Requires broader language-variation testing |

### 4.4 Planned ablation analysis

A formal ablation experiment has not yet been completed. The planned evaluation compares text-only, rule-only, text-plus-rule, voice-safety-only, full text-rule-voice fusion, and full Anian without generation gating. This design will quantify the independent contributions of rules, multimodal safety fusion, and generation gating to high-risk recall, false-positive burden, escalation accuracy, and unsafe generation exposure.

**Table 5 |** Safety-fusion ablation-study design.

| Experiment | System composition | Evaluation target | Key metric | Interpretive goal |
| --- | --- | --- | --- | --- |
| A | Text model only | Learning-model baseline | macro-F1; high-risk recall | Assess model limitations |
| B | Rule only | Safety-rule coverage | high-risk recall; FPR | Evaluate rule fallback and false positives |
| C | Text + rule | Rule enhancement | FNR reduction | Quantify safety value of L3 rules |

| D | Voice safety only | External speech-safety contribution | recall; FPR | Evaluate auxiliary speech service |
|---|---|---|---|---|
| E | Text + rule + voice | Full safety fusion | high-risk recall; escalation accuracy | Evaluate maximum-risk fusion |
| F | Full system without generation gate | Effect of removing generation control | unsafe generation exposure | Evaluate necessity of response gating |

### 4.5 Error-analysis framework

To avoid over-reliance on aggregate metrics, subsequent evaluation will include categorized error analysis. Target error categories include semantic mixture, negation handling, quotation or third-party context, ASR transcription error, multi-turn risk accumulation, and route-acceptability error.

**Table 6 |** Error-analysis framework.

| Error type | Typical scenario | Potential consequence | Improvement direction |
|---|---|---|---|
| Semantic mixture | Sleep distress, self-blame, and anxiety co-occur | Construct omissions or route shift | Multi-label and contextual modeling |
| Negation error | "I do not have suicidal thoughts" | Unnecessary escalation | Negation detection and gold-set testing |
| Quotation/context error | Discussion of news or another person’s suicide | False positive | Context and speaker-role modeling |
| ASR error | High-risk phrase lost under crying/noise/accent | Missed escalation | Speech robustness testing |
| Multi-turn risk accumulation | Risk emerges across several turns | Delayed escalation | Conversation-trajectory modeling |
| Routing ambiguity | Both breathing and sleep practice are reasonable | Overly strict top-1 error | Expert-defined acceptable-route range |

## 5 Discussion

### 5.1 Principal findings

This paper presents Anian as a safety-gated multimodal AI backend for mental-health support, with perinatal mindfulness routing as the motivating application. The principal contribution is architectural: ordinary generative interaction is downstream of structured state representation, conservative safety fusion, and a generation gate. The prototype technical evaluation suggests that the current implementation can learn the weak-label state framework, execute predefined safety rules, route low-risk support content, and block ordinary generation in designed moderate/high-risk scenarios.

The key design principle is that, in safety-sensitive mental-health contexts, intelligence is not only the ability to produce an appropriate sentence. It is also the ability to preserve high-risk evidence, select an appropriate route, and determine when not to generate.

### 5.2 Why conservative fusion matters

Multimodal systems may receive conflicting signals. A user may speak calmly while expressing self-harm intent in words, or an ASR transcript may be ambiguous while prosodic features suggest distress. Anian uses highest-risk-priority fusion because averaging or majority voting could downgrade rare but critical safety evidence. This approach may increase false positives, but it is intended to reduce high-risk false negatives.

This design also separates voice tone from safety adjudication. Tone and prosody can shape style, timing, or follow-up questions in low-risk settings, but they should not reduce risk when explicit high-risk text is present.

### 5.3 Relevance to perinatal support

Perinatal users may experience physiological changes, sleep deprivation, infant-care burden, role transition, and social pressure. Distress can move from ordinary anxiety or low mood to acute loss of control or safety risk. When infant safety is involved, the system must consider not only self-harm risk but also possible harm to an infant or others. This requires explicit escalation logic rather than only empathic conversation.

Anian treats mindfulness practices as routable support resources, not as a static content library. Worry can map to breathing or grounding, sleep distress to a short sleep practice, self-blame to self-compassion, isolation to support-seeking prompts, and moderate/high safety risk to safety escalation. This aligns personalization with safety-aware action selection.

### 5.4 Evidence boundaries

The current evidence supports internal technical feasibility only. The corpus is assembled from public emotion, dialogue, mental-health-related, and Chinese dialogue datasets; it is not a perinatal clinical interview gold-standard dataset. L2 and L4 labels are partly weakly supervised or rule-derived. The L3 safety result comes from a controlled stress test and should not be interpreted as real-world zero false negatives. ASR robustness, dialect/accent variation, multi-turn risk accumulation, user acceptability, clinician workflow integration, and safety-script review remain incomplete.

For these reasons, Anian should not be described as clinically validated, diagnostically accurate, or deployment-ready. The correct interpretation is that the architecture provides a testable safety-gated pathway that can be subjected to stronger human and real-world evaluation.

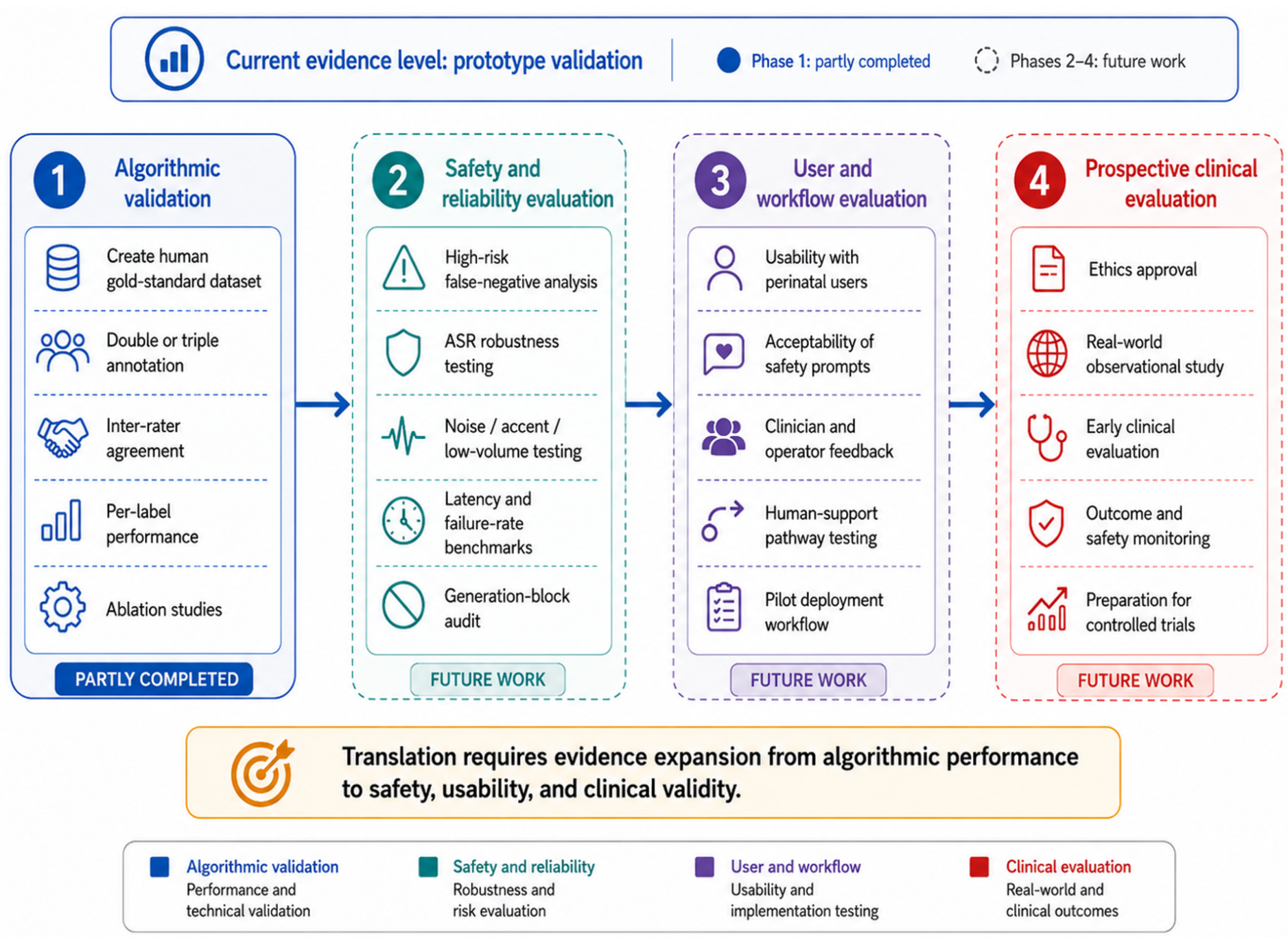

**Figure 4 |** Proposed validation roadmap from prototype technical evaluation to clinical translation. The current evidence level is internal prototype evaluation. Later phases require locked human gold-standard data, expert safety review, ASR robustness testing, ablation experiments, user and workflow studies, and prospective clinical evaluation.

**Table 7 |** Evidence maturity and validation roadmap.

| Evidence domain | Current status | Validation-phase enhancement | Expected output |
|---|---|---|---|
| Human gold standard | Not yet established | Double/triple annotation and expert adjudication | Locked gold set and kappa report |
| Expert safety review | Documentation incomplete | Review of L3, safety scripts, SOP, and course notices | Expert review forms and frozen versions |
| ASR robustness | Not yet systematically tested | Noise, accent, low-volume, and emotional-speech testing | CER and safety-recall tables |
| Ablation evidence | Formal ablation pending | Text-only/rule-only/voice/full comparisons | Ablation results table |
| Real-world usability | Not yet evaluated | Perinatal user and clinician workflow study | Acceptability and feasibility outcomes |

| Clinical validity | Not yet established | Prospective clinical evaluation | Outcome and safety monitoring data |
|---|---|---|---|

## 6 Limitations

First, the evaluation corpus was assembled primarily from public emotion, dialogue, mental-health-related, and Chinese dialogue datasets; it is not a perinatal clinical interview gold-standard dataset. Second, L2 and L4 labels are partly weakly supervised or rule-derived, so performance metrics mainly indicate learning of the current label framework rather than clinical construct validity. Third, the L3 safety result comes from a controlled stress test and should not be interpreted as real-world zero false negatives. Fourth, ASR errors, low-volume speech, crying or choked voice, background noise, accent variation, dialectal Mandarin, and abnormal speech rate have not yet been systematically tested. Fifth, the current evaluation does not include human gold-standard annotation, expert adjudication, prospective users, clinical outcomes, safety-event monitoring, or workflow integration. Sixth, the generation gate has been evaluated as an engineering control, but downstream fixed scripts, support prompts, and human-support pathways require independent expert review before real-world use.

## 7 Future work

The next phase will establish a perinatal Chinese human gold-standard dataset with independent annotation and expert adjudication. L1-L4 labels should be evaluated separately. L3 safety risk should be evaluated with special emphasis on high-risk recall, false-negative rate, false-positive rate, and categorized error examples. Speech robustness testing should report ASR character error rate together with safety recall. A formal ablation study should quantify the contributions of text models, rules, external voice safety, and generation gating. A later prospective study should evaluate usability, workflow integration, acceptability of safety prompts, human-support pathway reliability, and safety outcomes under ethical oversight.

## 8 Conclusion

Anian demonstrates a safety-gated multimodal AI backend for mental-health support and perinatal mindfulness-intervention routing. By combining hierarchical state representation, rule-based safety control, conservative multimodal fusion, and generation gating, the system provides a structured pathway from user expression to support routing and safety escalation. Its current evidence base supports internal prototype feasibility, not clinical effectiveness or deployment readiness. The most important next step is not merely to improve conversational fluency, but to establish human-validated, expert-reviewed, and real-world-tested evidence that the system can safely determine when to support, when to escalate, and when not to generate.

## Declarations

**Ethics statement:** This study used public datasets, internal engineering test cases, and non-clinical prototype evaluation. It did not recruit human participants, perform a prospective clinical intervention, or analyze identifiable patient-level outcome data. Ethics approval was therefore not required for the work reported here. Any subsequent study involving real users, clinical workflows, identifiable data, or intervention outcomes will require appropriate ethics review and oversight before initiation.

**Clinical trial registration:** Not applicable. This manuscript does not report a clinical trial.

**Data availability:** The public source datasets are available from their original providers subject to their respective licenses and terms of use. The current merged Anian corpus, rule mappings, generated weak labels, stress-test prompts, and internal preprocessing manifest are not publicly released in this version because they contain development-specific safety-trigger libraries and licensing-dependent processed data. A de-identified, license-compliant evaluation manifest should be prepared for future public release or independent audit.

**Code availability:** A de-identified description of the model pipeline, evaluation workflow, and rule categories is provided in the manuscript and appendix. Production deployment configuration, full safety-trigger libraries, server addresses, and internal orchestration code are not publicly released.

**Funding:** This internal prototype technical evaluation received no external funding.

**Competing interests:** Lei Wang is affiliated with RYTECH, which is involved in the development and potential commercialization of Anian. The authors declare no other competing interests.

**Author contributions:** Lei Wang and Xiao Wang equally contributed to conceptualization, system architecture, prototype implementation, methodology, evidence synthesis, manuscript drafting, and project supervision. Lei Li contributed to methodological interpretation, neuroscience and AI-safety framing, manuscript review, and critical revision. Both authors reviewed and approved the final manuscript.

**AI assistance disclosure:** General-purpose AI writing and editing tools were used to assist with language polishing, structural revision, and formatting. The authors reviewed, edited, and approved the content and are responsible for the accuracy, integrity, references, and conclusions of the manuscript.

# Appendix: Ontology, safety rules, and evaluation protocols

**Supplementary Table S1 |** L1 emotion-state ontology.

| Layer | Label | Operational definition | Example expression | Safety role | Potential route |
|---|---|---|---|---|---|
| L1 | neutral | Low signal or no clear emotional expression | "I see"; "It is okay" | Does not directly affect safety | light_check_in |
| L1 | sadness | Sadness, low mood, helplessness, crying-related expression | "I keep wanting to cry" | Supports low-mood detection | light_check_in / self_compassion |
| L1 | anxiety_fear | Worry, fear, tension, repetitive checking | "I keep worrying whether the baby is okay" | Signals anxiety burden | breathing_grounding |
| L1 | anger_irritability | Irritability, anger, easy escalation | "I explode at the smallest thing" | May indicate overload | downshift_grounding |
| L1 | guilt_shame | Guilt, shame, self-devaluation | "I am not a good mother" | Related to self_blame | self_compassion |
| L1 | calm_relief | Relief, calmness, easing | "This makes me feel a bit lighter" | May indicate settling | settling / light_check_in |
| L1 | joy_positive | Positive affect or hopefulness | "I feel better today" | Low-risk support | light_check_in |

**Supplementary Table S2 |** L2 psychosocial-construct ontology.

| Layer | Label | Operational definition | Example expression | Intervention meaning | Routing implication |
|---|---|---|---|---|---|
| L2 | worry | Repetitive worry, anticipatory anxiety, reassurance seeking | "I keep worrying about the next check-up" | Reduce arousal | breathing_grounding |
| L2 | sleep_distress | Insomnia, nighttime distress, fatigue | "I cannot sleep at night and keep thinking" | Sleep-oriented grounding | sleep_grounding_short_practice |
| L2 | self_blame | Self-blame, shame, not feeling good enough | "It is all my fault" | Self-compassion support | self_compassion |
| L2 | relationship_stress | Partner, family, or social-support strain | "No one understands me" | Support communication | support_seeking_prompt |
| L2 | caregiving_stress | Feeding, infant care, caregiving pressure | "I cannot take care of the baby well" | Caregiving-pressure downshift | downshift_grounding |
| L2 | isolation | Loneliness, unsupported, abandoned feeling | "I am carrying this alone" | Connection to support | support_seeking_prompt |
| L2 | emotional_overload | Overload, collapse, loss-of-control edge | "I cannot hold it anymore" | Downshift plus safety check | downshift_grounding / safety_escalation |
| L2 | help_seeking | Explicit request for help | "What should I do?" | Real support guidance | ask_support_or_professional_path |
| L2 | acute_or_safety_distress | Acute pressure or safety-related distress | "I am afraid I may do something" | Must be interpreted with L3 | safety_escalation |

**Supplementary Table S3 |** L3 safety-risk levels and system actions.

| Safety level | Operational definition | Representative expression | System action | Generative response | Human/real-world support |
|---|---|---|---|---|---|
| none | No safety signal identified | "I feel a bit tired" | Ordinary state recognition | Allowed | Not mandatory |
| low | Low-risk or background distress expression | "I feel down today" | Light check-in or practice | Allowed | May prompt support |
| moderate | Acute loss of control, ambiguous safety concern, or need for real support | "I feel I may lose control" | Safety-oriented response | Ordinary generation blocked | Recommend trusted/professional support |
| high | Direct self-harm, suicide, infant/others-harm expression | "I do not want to live"; "I fear hurting the baby" | safety_escalation | Ordinary generation blocked | Strong prompt for real-world support/emergency resources |

**Supplementary Table S4 |** L4 intervention-routing ontology.

| Route label | Definition | Trigger condition | Frontend action | Safety note |
|---|---|---|---|---|
| light_check_in | Low-risk supportive check-in | Neutral or mild low mood | Brief response and continued listening | No diagnosis |
| breathing_grounding | Breathing or grounding exercise | worry; anxiety_fear | Short breathing practice | Not a substitute for high-risk support |
| downshift_grounding | Downshift for overload | emotional_overload; anger | Slow-down grounding | Requires safety exclusion first |
| sleep_grounding_short_practice | Sleep-related short practice | sleep_distress | Sleep practice/audio | Escalate if acute loss of control |
| self_compassion | Self-compassion support | self_blame; guilt_shame | Self-compassion audio | No medical diagnosis |
| support_seeking_prompt | Prompt to seek nearby support | isolation; help_seeking | Contact trusted person | Escalate if moderate/high risk |
| ask_support_or_professional_path | Professional-support path prompt | Persistent distress or explicit | Doctor/counselor/hotline prompt | Not a replacement for care |

| | | help seeking | | |
|---|---|---|---|---|
| safety_escalation | Safety escalation | L3 moderate/high | Fixed safety script and support path | Blocks ordinary generation |

**Supplementary Table S5 |** Safety-rule taxonomy and representative handling.

| Rule category | Representative semantics | Risk tendency | Handling principle | Comment |
|---|---|---|---|---|
| Direct self-harm | Suicide, not wanting to live, self-injury | high | Force safety escalation | Highest priority |
| Infant/others harm | Fear of hurting baby or others | high | Force safety escalation | Perinatal-specific safety scenario |
| Acute loss of control | Collapse, cannot control self | moderate/high | Safety-oriented response | Context dependent |
| Implicit despair | Disappearing, life has no meaning | moderate | Prompt real support | Requires gold validation |
| Negation filter | No suicidal thoughts; will not hurt baby | low/none | Do not automatically escalate | May retain supportive check-in |
| Quotation/third-party | News or another person’s self-harm | low/none | Context filtering | Not user’s own risk |
| Colloquial false positives | Slang or idioms containing high-risk words | none/low | False-positive suppression | Important in Chinese context |

**Supplementary Table S6 |** Human annotation workflow.

| Step | Procedure | Output | Quality control |
|---|---|---|---|
| 1. Sampling | Stratified sampling from de-identified real, expert-simulated, and stress-test scenarios | Candidate pool | Balance ordinary, high-risk, and false-positive cases |
| 2. Manual | Define L1-L4 labels, examples, and exclusion rules | Annotation manual | Revise after pilot annotation |
| 3. Independent annotation | At least two raters annotate independently | Double-coded data | Hide model predictions |
| 4. Agreement analysis | Compute kappa/agreement | Agreement report | Report L3 separately |
| 5. Expert adjudication | Safety-related disagreements adjudicated by experts | Final gold labels | High-risk safety priority |
| 6. Test-set lock | Freeze version and hash | Human gold test set | Not reused for training |

**Supplementary Table S7 |** Gold-set evaluation metrics.

| Layer | Primary metric | Recommended reporting | Interpretive focus |
|---|---|---|---|
| L1 | Per-label precision/recall/F1 | Macro-F1 and label-wise F1 | Low-frequency emotion labels |
| L2 | Multi-label F1 | Micro/macro/weighted F1 | Non-diagnostic construct recognition |
| L3 | High-risk recall/FNR/FPR | Confusion matrix and error examples | Safety recall priority |
| L4 | Top-1 accuracy and acceptable-route agreement | Strict and acceptable routes | Multiple reasonable interventions |

**Supplementary Table S8 |** ASR robustness testing protocol.

| Speech condition | Sample type | Primary metric | Risk concern | Reporting format |
|---|---|---|---|---|
| Clean Mandarin | Low/moderate/high-risk utterances | CER; safety recall | Baseline | Main table |
| Low volume | Fatigued or subdued voice | CER; FNR | Loss of key safety words | Subgroup table |
| Crying/choked voice | High emotional load | CER; escalation recall | Crisis-like speech | Error analysis |
| Background noise | Infant crying, home, traffic | CER; FPR/FNR | Home-use condition | Robustness figure |
| Accent/dialectal Mandarin | Regional accents | CER; route shift | Robustness and fairness | Stratified results |
| Abnormal speech rate | Fast or slow speech | CER; tone consistency | Tone shapes style but not risk | Fusion audit |

**Supplementary Table S9 |** Ablation study design.

| Version | Composition | Purpose | Key metric | Use |
|---|---|---|---|---|
| A | Text model only | Learning-model baseline | macro-F1; high-risk recall | Evaluate model alone |
| B | Rule only | Safety-rule capability | FNR; FPR | Define rule boundary |
| C | Text + rule | Rule enhancement | FNR reduction | Demonstrate L3 necessity |
| D | Voice safety only | External voice contribution | recall; FPR | Evaluate speech service |
| E | Full Anian | Full fusion | high-risk recall; escalation accuracy | Primary system result |

| F | Full without gate | Remove generation gate | unsafe generation exposure | Demonstrate gate necessity |
|---|---|---|---|---|

**Supplementary Table S10 |** Generation-gate audit metrics.

| **Audit item** | **Definition** | **Calculation** | **Interpretation** |
|---|---|---|---|
| Block rate | Proportion of samples for which generation is blocked | blocked / total | System conservativeness |
| Appropriate block rate | True moderate/high-risk samples correctly blocked | correct blocks / true moderate-high | Safety recall |
| Over-block rate | Low-risk samples incorrectly blocked | incorrect blocks / true low | User-experience burden |
| Leakage rate | Moderate/high-risk samples not blocked | missed blocks / true moderate-high | Critical safety risk |
| Post-block action accuracy | Correct downstream support action after block | correct action / blocked | Closed-loop quality |

**Supplementary Table S11 |** Expert safety review matrix.

| **Review target** | **Reviewer expertise** | **Review dimension** | **Record** | **Version control** |
|---|---|---|---|---|
| L3 safety-level definitions | Psychiatry/crisis-intervention expert | Risk boundary and miss risk | Pass/revise/reject | Version and date |
| Fixed safety scripts | Crisis-intervention and compliance expert | Clarity, non-diagnostic wording, real-world support direction | Reviewer form | Frozen script version |
| Mindfulness-course notices | Perinatal clinician or psychologist | Appropriate boundary between support and treatment | Reviewer form | Course version and date |
| Human-support pathway | Clinical workflow or operations expert | Escalation feasibility and local-resource wording | Reviewer form | SOP version |
| Generation-gate policy | AI safety and compliance expert | Leakage risk, over-block burden, auditability | Gate audit record | Model/rule version |
| Speech/ASR safety protocol | Speech technology and clinical-safety expert | Robustness under noise, accent, low volume, and emotional speech | Robustness report | ASR/service version |